\documentclass[10pt]{article}

\usepackage[utf8]{inputenc}
\usepackage[T1]{fontenc}
\usepackage[pdftex]{graphicx}
\graphicspath{{./pdf/}}

\usepackage{csagh}
\usepackage{booktabs}
\usepackage{multirow}
\usepackage{pgfplots}
\pgfplotsset{compat=1.17}

\begin{document}
\begin{opening}

\title{Bielik Guard: Efficient Polish Language Safety Classifiers for LLM Content Moderation}

\author[SpeakLeash Foundation, Warsaw, Poland, krzysztof.wrobel@bielik.ai \\
        Jagiellonian University, Cracow, Poland, krzysztof.pawel.wrobel@uj.edu.pl]{Krzysztof Wróbel}

\author[SpeakLeash Foundation, Warsaw, Poland, jan.maria.kowalski@bielik.ai]{Jan Maria Kowalski}

\author[Warsaw School of Economics, Warsaw, Poland, jerzy.surma@sgh.waw.pl]{Jerzy Surma}

\author[SpeakLeash Foundation, Warsaw, Poland, igor.ciuciura@bielik.ai]{Igor Ciuciura}

\author[SpeakLeash Foundation, Warsaw, Poland, maciej.szymanski@bielik.ai]{Maciej Szymański}

\begin{abstract}
As Large Language Models (LLMs) become increasingly deployed in Polish language applications, the need for efficient and accurate content safety classifiers has become paramount. We present Bielik Guard, a family of compact Polish language safety classifiers comprising two model variants: a 0.1B parameter model based on MMLW-RoBERTa-base and a 0.5B parameter model based on PKOBP/polish-roberta-8k. Fine-tuned on a community-annotated dataset of 6,885 Polish texts, these models classify content across five safety categories: Hate/Aggression, Vulgarities, Sexual Content, Crime, and Self-Harm. Our evaluation demonstrates that both models achieve strong performance on multiple benchmarks. The 0.5B variant offers the best overall discrimination capability with F1 scores of 0.791 (micro) and 0.785 (macro) on the test set, while the 0.1B variant demonstrates exceptional efficiency. Notably, Bielik Guard 0.1B v1.1 achieves superior precision (77.65\%) and very low false positive rate (0.63\%) on real user prompts, outperforming HerBERT-PL-Guard (31.55\% precision, 4.70\% FPR) despite identical model size. The models are publicly available and designed to provide appropriate responses rather than simple content blocking, particularly for sensitive categories like self-harm.
\end{abstract}


\end{opening}

\section{Introduction}

The rapid advancement of Large Language Models has revolutionized natural language processing capabilities, enabling sophisticated conversational AI systems across numerous domains \cite{dubey2024llama3}. However, this progress brings significant challenges in ensuring safe and responsible deployment, particularly in multilingual contexts where safety resources remain scarce \cite{inan2023llamaguard}. 

For Polish language applications, the landscape of LLM safety tools has been particularly limited. Existing solutions either rely on English-centric models adapted to Polish with varying degrees of success, or employ large multilingual models that may be impractical for resource-constrained deployments. The need for dedicated Polish safety classifiers is further motivated by cultural and linguistic nuances that affect what constitutes harmful content and how it should be moderated.

We introduce Bielik Guard (codenamed Sójka, meaning ``jay'' in Polish -- a vigilant bird symbolizing protection), a family of efficient safety classifiers specifically designed for Polish language content. Our contributions include:

\begin{itemize}
    \item Two compact model variants (0.1B based on MMLW-RoBERTa and 0.5B based on PKOBP/polish-roberta-8k) \cite{dadas2026longcontextencodermodelspolish} optimized for deployment efficiency while maintaining high accuracy
    \item A community-driven annotation methodology based on bounded rationality principles \cite{simon1982satisficing}, yielding 6,885 annotated Polish texts with over 60,000 individual ratings
    \item A five-category safety taxonomy tailored to Polish language applications: Hate/Aggression, Vulgarities, Sexual Content, Crime, and Self-Harm
    \item Comprehensive evaluation demonstrating superior precision and lower false positive rates compared to larger multilingual alternatives
    \item A response-oriented approach that provides appropriate support resources rather than simple blocking, especially for self-harm content
\end{itemize}

The models are publicly available at \url{https://huggingface.co/speakleash} and have been deployed in production at \url{https://guard.bielik.ai/}, where ongoing community feedback continues to improve the system.

\section{Related Work}

\subsection{LLM-based Safety Classifiers}

The development of safety guardrails for LLMs has become a critical research area, with comprehensive surveys covering the current state of the art \cite{dong2024safeguardinglargelanguagemodels,ayyamperumal2024currentstatellmrisks,ZHANG2025100301}. Llama Guard \cite{inan2023llamaguard} pioneered the approach of using instruction-tuned language models for input-output safety classification, introducing a taxonomy-based framework that allows adaptation to different use cases. The subsequent Llama Guard 3 \cite{dubey2024llama3} extended this work with an 8B parameter model supporting multilingual classification across 14 MLCommons hazard categories, achieving F1 scores of 0.939 on English response classification.

Similarly, Qwen3Guard \cite{qwen3guard} introduced three-tiered severity classification (safe, controversial, unsafe) with support for 119 languages, offering models ranging from 0.6B to 8B parameters. ShieldGemma \cite{zeng2024shieldgemmagenerativeaicontent}, based on the Gemma model family, provides content moderation capabilities with models of various sizes designed for different deployment scenarios. Granite Guardian \cite{padhi2024graniteguardian} extends beyond traditional harmful-content detection by unifying prompt and response risk detection with coverage of social bias, profanity, violence, sexual content, unethical behavior, jailbreaking, and RAG-specific hallucination risks (context relevance, groundedness, answer relevance). The 2B and 8B variants, trained on a combination of human-annotated and synthetic data, achieve AUC scores of 0.871 and 0.854 on harmful-content and RAG groundedness benchmarks respectively. These generative models frame safety as an instruction-following task, enabling flexible deployment scenarios.

\subsection{Polish Language Models and Safety}

The development of Polish language models has accelerated in recent years. The Bielik family of models \cite{Ociepa_Flis_Wróbel_Gwoździej_Kinas_2025,ociepa2025bielik11bv2technical,ociepa2025bielikv3smalltechnical,ociepa2025bielik11bv3technical} and the PLLuM family \cite{kocon2025pllumfamilypolishlarge} represent significant milestones in Polish LLM development, demonstrating the growing maturity of Polish NLP infrastructure. These foundational models highlight the importance of dedicated Polish language resources and the need for corresponding safety mechanisms.

For Polish-specific safety classification, HerBERT-PL-Guard \cite{plguard2025} represents a significant contribution, fine-tuning the HerBERT model on manually annotated data and Polish translations of PolyGuard and WildGuard datasets. The model supports classification into 15 categories based on the Llama Guard taxonomy, including both safe and 14 unsafe categories.

However, existing Polish solutions face limitations: HerBERT-PL-Guard, while comprehensive in its taxonomy, is trained on a mixture of manually annotated Polish data and translated datasets (PolyGuard/WildGuard), which may not fully reflect the distribution and idiomatic patterns of real-world Polish user traffic. Additionally, its artifacts are released under the CC BY-NC-SA 4.0 license, which restricts commercial use and may complicate adoption in production deployments. Multilingual models like Llama Guard 3, despite their broad language coverage, exhibit higher false positive rates and lower precision on Polish content, as our evaluation demonstrates.

\subsection{Community-based Annotation}

Our approach to data collection draws inspiration from crowdsourcing methodologies in NLP while incorporating principles from bounded rationality \cite{simon1982satisficing}. Rather than seeking an objective ground truth, we embrace the notion that safety judgments are inherently subjective and context-dependent, making community consensus a more appropriate target than expert-only annotation.

A key design assumption was that disagreements between annotators are not noise but an informative signal reflecting ambiguity, cultural context, and individual moral intuition. This is particularly relevant in the Polish language, where slang, idiomatic expressions, and pragmatic meanings often blur the boundary between harmless and harmful intent.

To operationalize this assumption, we designed a lightweight, purpose-built annotation platform that allowed volunteers to label short text samples through a simple survey interface. Texts were randomly assigned to annotators to minimize ordering effects and individual bias. A visible counter of completed annotations was deliberately introduced as a motivational mechanism, which proved effective in sustaining engagement during the early, high-volume phase of the campaign.

Community engagement was not limited to a one-off data collection effort. Instead, it continues through an ongoing public annotation interface available at \url{https://guard.bielik.ai/ankieta.html}, enabling iterative dataset expansion and future recalibration of the model as social norms evolve.

The annotation campaign was promoted through webinars, social media channels, and community-driven outreach around the Bielik ecosystem. This resulted in rapid scaling: over 25,000 annotations were submitted within the first week alone, demonstrating both the accessibility of the interface and the willingness of non-expert users to participate in AI safety–oriented initiatives.

\section{Bielik Guard: Model Architecture and Training}

\subsection{Safety Taxonomy}

Bielik Guard employs a five-category taxonomy designed specifically for Polish language safety needs:

\begin{itemize}
    \item \textbf{HATE (Hate/Aggression):} Content attacking or discriminating against groups based on race, religion, gender, sexual orientation, or nationality
    \item \textbf{VULGAR (Vulgarities):} Profane or vulgar language in both explicit and masked forms
    \item \textbf{SEX (Sexual Content):} Graphic descriptions of sexual activities or requests for erotic material generation
    \item \textbf{CRIME:} Instructions or encouragement for criminal activities including drug production and fraud
    \item \textbf{SELF-HARM:} Content encouraging suicide, self-harm, or eating disorders
\end{itemize}

This taxonomy deliberately excludes categories like disinformation, jailbreaking attempts, and copyright violations for several reasons: (1) these categories require factual knowledge that may change over time, (2) detecting such content often requires context beyond isolated text snippets, and (3) our community annotation process focused on immediate safety risks that require active intervention or support. This focused scope allows for more consistent annotation and clearer deployment guidelines.

\subsection{Model Architecture}

The two Bielik Guard variants employ different base models, chosen to explore different points in the efficiency-performance trade-off space:

\textbf{Bielik Guard 0.1B} is built upon MMLW-RoBERTa-base \cite{sdadas2020mmlw}, a 124M parameter Polish RoBERTa-based encoder \cite{liu2019roberta} with a vocabulary of 50,001 tokens, producing 768-dimensional representations.

\textbf{Bielik Guard 0.5B} is built upon PKOBP/polish-roberta-8k \cite{pkobp2023roberta}, a 443M parameter Polish RoBERTa variant with an enhanced vocabulary of 128,064 tokens, providing substantially greater modeling capacity.

For both variants, we add a multi-label classification head \cite{zhang2014review} consisting of:

\begin{itemize}
    \item A dropout layer (p=0.1) for regularization
    \item A linear projection layer mapping hidden dimensions to 5 output logits
    \item Sigmoid activation for independent binary classification per category
\end{itemize}

\subsection{Training Data and Methodology}
\subsubsection{Data Collection}

\label{sec:data-collection}

The training dataset comprises 6,885 unique Polish texts collected through large-scale community engagement:

\begin{itemize}
\item Over 1,500 volunteers participated in annotation
\item Each text received an average of 7--8 independent ratings
\item A total of over 60,000 individual annotations were collected
\item Sources included anonymized user prompts from Polish LLM interactions as well as selected social media content
\end{itemize}

The resulting dataset exhibits a relatively balanced distribution, with approximately 55\% of samples labeled as safe and 45\% as harmful or potentially unsafe. This balance was achieved deliberately to avoid over-representation of benign content while preserving the prevalence of borderline and controversial cases observed in real-world usage.

Rather than binarizing annotations, we trained the model on the percentage of annotators who classified each text as belonging to a given category. This regression-based labeling strategy preserves information about annotation agreement and explicitly models controversial cases.

For example, a text labeled as \textit{HATE} by 66\% of annotators is treated differently from one labeled unanimously, allowing the model to learn graded risk signals instead of hard thresholds. This approach avoids premature discretization decisions (e.g., 50\%, 3/5, or 4/6 majority rules) and defers the choice of decision boundaries to downstream applications.

\subsubsection{Training Splits and Evaluation Strategy}

We employed two training configurations to balance comprehensive evaluation with maximal production performance:

\textbf{Configuration 1: 2:1 Split (2,295 train / 4,590 test).} To enable statistically robust evaluation with a large test set, we initially trained models on a 2:1 split of the dataset. This configuration prioritizes having sufficient test samples for comprehensive analysis. Results on the Sojka test set and Sojka augmented test set (Tables \ref{tab:sojka-results} and \ref{tab:augmented-results}) use models trained with this split.

\textbf{Configuration 2: Near-Complete Training (6,285 train / 600 test).} To maximize production performance, we trained models on the near-complete dataset, using almost all available data for training. Results on the Gadzi Jezyk benchmark (Table \ref{tab:gadzi-results}) and comparison with state-of-the-art models on user prompts (Table \ref{tab:comparison}) use models trained with this configuration, as they represent our best-performing models for deployment.

The relatively small training set size in Configuration 1 is offset by the strong linguistic representations already learned by the base models (MMLW-RoBERTa-base and PKOBP/polish-roberta-8k). Multi-label classification is fully supported, allowing texts to belong to multiple categories simultaneously.

\subsubsection{Data Distribution}

We consider a text to belong to a category if at least 60\% of annotators classified it as such. While model training uses continuous percentage values (0-100\%) as soft labels, evaluation metrics requiring binary ground truth (F1, precision, recall, specificity) use this 60\% threshold for binarization. Model predictions are binarized at the standard 0.5 threshold. Using this 60\% agreement threshold, the dataset shows natural class imbalance:

\begin{itemize}
    \item Safe content: 3,781 texts (54.92\%)
    \item SELF-HARM: 796 texts (11.56\%)
    \item HATE: 988 texts (14.35\%)
    \item SEX: 895 texts (13.00\%)
    \item VULGAR: 411 texts (5.97\%)
    \item CRIME: 311 texts (4.52\%)
\end{itemize}

\subsubsection{Quality Control}

Quality assurance included deduplication, clustering analysis to verify annotation consistency, and expert validation for controversial cases. The methodology explicitly embraces bounded rationality, targeting satisficing rather than optimal solutions and treating safety as a matter of social consensus rather than objective truth.

We deliberately do not report traditional inter-annotator agreement metrics as these assume the existence of a single "correct" label. Instead, our soft-label approach treats disagreement as informative signal about the inherent ambiguity and context-dependence of safety judgments. The variance in annotation percentages naturally captures the degree of consensus: texts with near-unanimous ratings (close to 0\% or 100\%) represent clear cases, while those with intermediate percentages (e.g., 40-60\%) reflect genuine ambiguity that the model learns to recognize.

\subsection{Training Procedure}

Models were fine-tuned using standard practices for transformer-based classification:

\begin{itemize}
    \item Loss function: Binary Cross-Entropy (BCE) with soft labels derived from percentage-based annotations. We experimented with Mean Squared Error (MSE) loss but found BCE to yield superior performance. No class weighting was applied.
    \item Optimizer: AdamW with weight decay of 0.01
    \item Learning rate: 2e-5 with 500 warmup steps followed by linear decay
    \item Batch size: 32
    \item Training duration: 3 epochs (approx. 2 hours on A100)
    \item Training infrastructure: A100 GPU cluster (ACK Cyfronet AGH)
\end{itemize}

The use of soft labels (annotation percentages) rather than hard binary labels allows the model to learn the degree of consensus among annotators, preserving information about controversial or ambiguous cases. For evaluation, ground truth labels are binarized at 60\% annotator agreement (reflecting majority consensus), while model predictions are binarized at the standard 0.5 sigmoid threshold. Users can adjust this prediction threshold based on their specific precision-recall requirements.

Both models were trained using the same training procedure and augmentation strategy. For the initial 2:1 split training, the test set was augmented using 15 text augmentation techniques (including diacritic manipulation, capitalization changes, character swaps, and spacing modifications) to evaluate model robustness.

\subsection{Model Versions}

Two versions of each model variant were developed:

\textbf{v1.0:} Initial models exhibiting overreaction to crime-related content due to a classification threshold calibration issue.

\textbf{v1.1:} Improved models with the crime category threshold issue resolved, resulting in substantially improved precision (77.65\% vs. 67.27\% for the 0.1B variant on user prompts, Table \ref{tab:comparison}) and lower false positive rates (0.63\% vs. 1.20\%). Both v1.0 and v1.1 models were trained using identical procedures and data splits; the difference lies solely in the threshold calibration fix.

For each training configuration (2:1 split and near-complete), we trained both v1.0 and v1.1 versions. To distinguish between configurations, we use the following versioning scheme:
\begin{itemize}
    \item \textbf{v1.0a / v1.1a:} Models trained with Configuration 1 (2:1 split: 2,295 train / 4,590 test)
    \item \textbf{v1.0 / v1.1:} Models trained with Configuration 2 (near-complete: 6,285 train / 600 test)
\end{itemize}

All subsequent analyses prioritize v1.1 models as they represent the production-ready variants with optimal precision-recall trade-offs.

\section{Evaluation}

We evaluate Bielik Guard on three datasets using metrics appropriate for multi-label classification: RMSE, F1 (micro and macro), Specificity, and ROC AUC. Additionally, we compare against state-of-the-art alternatives on user prompt data to assess practical deployment performance.

\subsection{Sojka Test Set}

The primary evaluation uses the held-out Sojka test set (4,590 samples) from the 2:1 split training configuration (Configuration 1), with the same label distribution as the full dataset. Results are shown in Table \ref{tab:sojka-results}.

\begin{table}[!ht]
\centering
\caption{Performance on Sojka test set. Ground truth binarized at 60\% annotator agreement; predictions binarized at 0.5 threshold.}
\label{tab:sojka-results}
\begin{tabular}{lcccc}
\toprule
\textbf{Metric} & \textbf{0.1B v1.0a} & \textbf{0.5B v1.0a} & \textbf{0.1B v1.1a} & \textbf{0.5B v1.1a} \\
\midrule
RMSE & 0.137 & 0.130 & 0.128 & \textbf{0.122} \\
F1 micro & 0.756 & 0.781 & 0.775 & \textbf{0.791} \\
F1 macro & 0.747 & 0.774 & 0.770 & \textbf{0.785} \\
Recall micro & 0.813 & \textbf{0.851} & 0.808 & 0.835 \\
Recall macro & 0.799 & \textbf{0.829} & 0.794 & 0.812 \\
Specificity micro & 0.961 & 0.962 & \textbf{0.968} & \textbf{0.968} \\
Specificity macro & 0.960 & 0.961 & \textbf{0.967} & \textbf{0.967}  \\
ROC AUC micro & 0.974 & 0.979 & 0.974 & \textbf{0.980}  \\
ROC AUC macro & 0.964 & \textbf{0.973} & 0.964 & \textbf{0.973}  \\
\bottomrule
\end{tabular}
\end{table}

The 0.5B v1.1a model demonstrates the best overall performance with F1 micro of 0.791 and F1 macro of 0.785, indicating superior discrimination capability. Both v1.1a models maintain high specificity (>0.96), demonstrating strong ability to correctly identify safe content.

\subsubsection{Per-Category Analysis}

Table \ref{tab:per-category} presents detailed per-category performance metrics for both model variants, revealing category-specific strengths and challenges.

\begin{table}[!ht]
\centering
\caption{Per-category performance breakdown on Sojka test set (v1.1a models)}
\label{tab:per-category}
\small
\begin{tabular}{lcccc}
\toprule
\textbf{Category} & \multicolumn{2}{c}{\textbf{0.1B v1.1a}} & \multicolumn{2}{c}{\textbf{0.5B v1.1a}} \\
\cmidrule(lr){2-3} \cmidrule(lr){4-5}
& \textbf{F1} & \textbf{ROC AUC} & \textbf{F1} & \textbf{ROC AUC} \\
\midrule
SELF-HARM & \textbf{0.886} & 0.991 & 0.879 & \textbf{0.992} \\
HATE & 0.628 & 0.919 & \textbf{0.667} & \textbf{0.934} \\
VULGAR & 0.742 & 0.973 & \textbf{0.750} & \textbf{0.977} \\
SEX & 0.889 & 0.988 & \textbf{0.915} & \textbf{0.993} \\
CRIME & 0.707 & 0.949 & \textbf{0.716} & \textbf{0.971} \\
\bottomrule
\end{tabular}
\end{table}

The HATE category presents the greatest challenge for both models (F1 of 0.628 and 0.667), likely due to the inherent subjectivity and context-dependence of hate speech annotation. CRIME also proves challenging (F1 of 0.707 and 0.716), which may reflect its lower prevalence in the training data (4.52\% of samples). The SELF-HARM and SEX categories achieve the strongest performance, with F1 scores exceeding 0.87 for both model variants. All categories maintain ROC AUC scores above 0.91, indicating consistent discriminative ability across the taxonomy.

\subsection{Robustness to Text Perturbations}

To evaluate robustness against adversarial modifications and natural text variations, we tested on the Sojka Augmented dataset, which applies 15 augmentation techniques including diacritic manipulation, capitalization changes, character-level perturbations, and spacing modifications. Results are in Table \ref{tab:augmented-results}. These results use models trained with Configuration 1 (2:1 split).

\begin{table}[!ht]
\centering
\caption{Performance on Sojka augmented test set}
\label{tab:augmented-results}
\begin{tabular}{lcccc}
\toprule
\textbf{Metric} & \textbf{0.1B v1.0a} & \textbf{0.5B v1.0a} & \textbf{0.1B v1.1a} & \textbf{0.5B v1.1a} \\
\midrule
RMSE & 0.183 & \textbf{0.167} & 0.181 & \textbf{0.163} \\
F1 micro & 0.632 & 0.683 & 0.638 & \textbf{0.694} \\
F1 macro & 0.615 & 0.660 & 0.619 & \textbf{0.679} \\
Recall micro & 0.606 & 0.675 & 0.621 & \textbf{0.686} \\
Recall macro & 0.585 & 0.642 & 0.602 & \textbf{0.650} \\
Specificity micro & 0.964 & 0.965 & 0.962 & \textbf{0.966} \\
Specificity macro & 0.963 & 0.964 & 0.961 & \textbf{0.965} \\
ROC AUC micro & 0.908 & \textbf{0.936} & 0.909 & 0.934 \\
ROC AUC macro & 0.880 & 0.913 & 0.884 & \textbf{0.915} \\
\bottomrule
\end{tabular}
\end{table}

While performance degrades on perturbed text as expected, the 0.5B v1.1a model shows substantially better robustness, with F1 micro of 0.694 compared to 0.638 for the 0.1B v1.1a model. This validates the effectiveness of combining a more capable base model (443M vs. 124M parameters, larger vocabulary) with augmentation-enhanced training.

\subsection{Gadzi Jezyk Benchmark}

We evaluated on the Gadzi Jezyk dataset \cite{gadzijezyk2024}, a challenging benchmark containing 520 toxic prompts with extreme class imbalance: 505 crime-related examples (97.1\%), 43 hate/violence (8.3\%), 31 self-harm (6.0\%), 18 sexual content (3.5\%), and 4 vulgarities (0.8\%). This distribution makes it particularly suitable for evaluating crime category performance. The Bielik Guard models (Sójka) evaluated here were trained using Configuration 2 (near-complete data: 6,285 train / 600 test), representing our best-performing models for deployment. Table \ref{tab:gadzi-results} shows results.

\begin{table}[!ht]
\centering
\caption{Performance on Gadzi Jezyk dataset (97.1\% crime-related content)}
\label{tab:gadzi-results}
\begin{tabular}{lcccc}
\toprule
\textbf{Metric} & \textbf{0.1B v1.0} & \textbf{0.5B v1.0} & \textbf{0.1B v1.1} & \textbf{0.5B v1.1} \\
\midrule
RMSE & 0.236 & \textbf{0.217} & 0.286 & 0.241 \\
Precision & 0.977 & 0.974 & \textbf{0.985} & 0.973 \\
Recall & 0.702 & \textbf{0.762} & 0.557 & 0.714 \\
F1 & 0.817 & \textbf{0.855} & 0.712 & 0.823 \\
Specificity & 0.995 & 0.994 & \textbf{0.998} & 0.994 \\
ROC AUC & 0.974 & \textbf{0.980} & 0.959 & 0.967 \\
\bottomrule
\end{tabular}
\end{table}

The Gadzi Jezyk dataset presents a particularly revealing evaluation scenario: with 97.1\% crime-related content, it directly tests the impact of our v1.0 to v1.1 threshold calibration fix. The results clearly demonstrate the precision-recall trade-off inherent in the calibration. The v1.1 models achieve slightly higher precision (98.5\% vs. 97.7\% for 0.1B, maintaining 97.3\% for 0.5B) and improved specificity (99.8\% vs. 99.5\% for 0.1B), while recall decreases substantially (55.7\% vs. 70.2\% for 0.1B, 71.4\% vs. 76.2\% for 0.5B).

Notably, the 0.5B v1.1 model maintains strong overall performance with F1 of 0.823 (vs. 0.855 for v1.0), representing only a 4\% reduction despite the significant recall decrease. This demonstrates effective precision-recall balancing. The 0.1B v1.1 shows a larger F1 reduction (0.712 vs. 0.817), reflecting its more conservative threshold calibration. The maintained high ROC AUC scores (95.9-98.0\%) across all variants demonstrate that underlying model discrimination capability remains excellent; the threshold adjustment shifts the operating point toward higher precision rather than degrading model quality.

This trade-off yields substantial benefits in production deployment. While v1.1 shows 4-15 percentage point recall reductions on this crime-saturated benchmark (the category where threshold calibration was specifically applied), it achieves 0.63\%-0.73\% FPR on diverse real user prompts (Table \ref{tab:comparison}), representing a 6-7× improvement over models with aggressive thresholds like HerBERT-PL-Guard (4.70\% FPR). The precision improvements on Gadzi Jezyk, though modest (0.8-1.1 percentage points), translate to dramatically lower false positive rates on real-world content where class distribution is balanced rather than 97\% toxic. For sustainable production deployment, this trade-off prioritizes user trust through high precision over maximum recall on synthetic adversarial benchmarks.

\subsection{Comparison with State-of-the-Art Models}
\label{sec:comparison}

To assess practical deployment performance, we evaluated Bielik Guard against HerBERT-PL-Guard \cite{plguard2025}, Llama Guard 3 (1B and 8B variants) \cite{dubey2024llama3}, and Qwen3Guard-Gen-0.6B \cite{qwen3guard} on 3,000 random user prompts collected from Polish LLM interactions. This evaluation dataset is distinct from the training dataset described in Section \ref{sec:data-collection} in two important ways: (1) it consists of randomly sampled user prompts without any prefiltering, whereas the Sojka training dataset was prefiltered to contain dangerous categories, and (2) it was gathered specifically for comparative evaluation purposes and was not used during model training. Results for Bielik Guard are from models trained with Configuration 2 (near-complete data: 6,285 train / 600 test).

\textbf{Evaluation Methodology:} All models were evaluated using their default thresholds and taxonomies. Critically, for practical deployment assessment, we evaluated each model on a \textbf{binary safe/unsafe basis}: any text flagged as unsafe in any category by a model was considered an "unsafe" prediction, regardless of the number or type of categories triggered. This binary evaluation approach ensures fair comparison across different taxonomies (Bielik Guard: 5 categories, HerBERT-PL-Guard: 15 categories, Llama Guard 3: 14 categories, Qwen3Guard-Gen: 9 categories), as the number of categories does not influence the fundamental question of whether content should be moderated. This metric directly reflects production deployment performance where the primary decision is whether to intervene, regardless of specific categorization.

The annotation protocol involved a two-annotator plus super-annotator scheme: each text flagged as unsafe by any classifier was independently annotated by two annotators, with a third super-annotator resolving disagreements. Crucially, annotation was performed separately for each model's taxonomy: annotators judged whether a text was unsafe according to the specific safety categories defined by Bielik Guard, Llama Guard, Qwen3Guard, and HerBERT-PL-Guard respectively. This ensures that each model's precision is measured against ground truth that faithfully reflects what that model's taxonomy is designed to detect. Due to resource constraints, we annotated only texts flagged by at least one classifier rather than the entire dataset, which precludes calculating recall metrics (as we lack ground truth for texts classified as safe by all models). This methodology focuses evaluation on precision and false positive rate, which are critical metrics for production deployment where excessive false positives harm user experience.

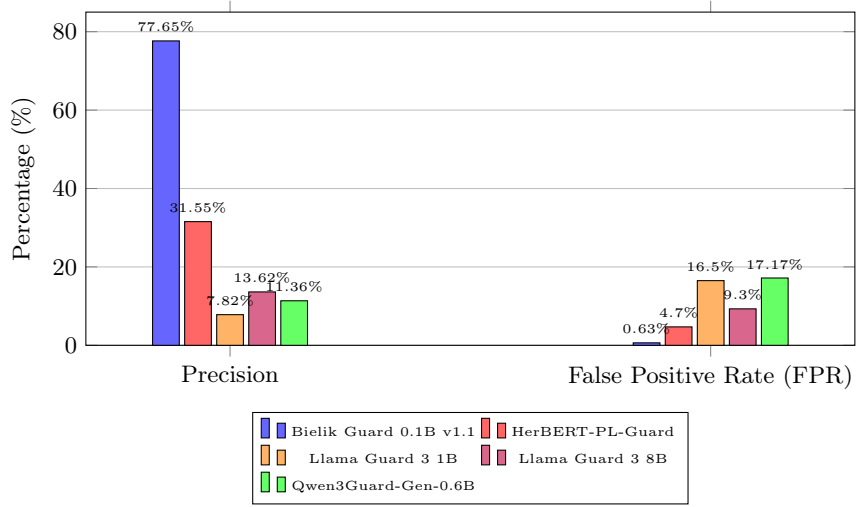
\begin{figure}[!ht]
\centering
\begin{tikzpicture}
\begin{axis}[
    ybar,
    bar width=10pt,
    width=0.9\textwidth,
    height=6cm,
    enlarge x limits=0.3,
    symbolic x coords={Precision, False Positive Rate (FPR)},
    xtick=data,
    nodes near coords={\pgfmathprintnumber\pgfplotspointmeta\%},
    nodes near coords style={font=\tiny},
    ymin=0, ymax=85,
    ylabel={Percentage (\%)},
    legend style={at={(0.5,-0.20)}, anchor=north, legend columns=2, font=\tiny},
    ymajorgrids=true,
]
\addplot[fill=blue!60] coordinates {(Precision,77.65) (False Positive Rate (FPR),0.63)};
\addplot[fill=red!60] coordinates {(Precision,31.55) (False Positive Rate (FPR),4.70)};
\addplot[fill=orange!60] coordinates {(Precision,7.82) (False Positive Rate (FPR),16.50)};
\addplot[fill=purple!60] coordinates {(Precision,13.62) (False Positive Rate (FPR),9.30)};
\addplot[fill=green!60] coordinates {(Precision,11.36) (False Positive Rate (FPR),17.17)};
\legend{Bielik Guard 0.1B v1.1, HerBERT-PL-Guard, Llama Guard 3 1B, Llama Guard 3 8B, Qwen3Guard-Gen-0.6B}
\end{axis}
\end{tikzpicture}
\caption{Comparison of safety classifiers on Polish user prompts. Higher Precision is better, lower FPR is better. Bielik Guard 0.1B v1.1 (124M) outperforms all compared models including larger multilingual alternatives.}
\label{fig:comparison}
\end{figure}

Results are presented in Table \ref{tab:comparison} and visualized in Figure \ref{fig:comparison}.

\begin{table}[!ht]
\centering
\caption{Comparison on 3,000 Polish user prompts (default thresholds). Each model evaluated with its own taxonomy. Recall not reported as only classifier-flagged texts were annotated.}
\label{tab:comparison}
\begin{tabular}{lcccc}
\toprule
\textbf{Model} & \textbf{Params} & \textbf{Precision} & \textbf{Alert Rate} & \textbf{FPR} \\
\midrule
Bielik Guard 0.1B v1.1 & 124M & \textbf{77.65\%} & \textbf{2.83\%} & \textbf{0.63\%} \\
Bielik Guard 0.5B v1.1 & 443M & \textbf{75.28\%} & \textbf{2.97\%} & \textbf{0.73\%} \\
Bielik Guard 0.1B v1.0 & 124M & \textbf{67.27\%} & \textbf{3.67\%} & \textbf{1.20\%} \\
HerBERT-PL-Guard & 124M & 31.55\% & 6.87\% & 4.70\% \\
Llama Guard 3 1B & 1B & 7.82\% & 17.90\% & 16.50\% \\
Llama Guard 3 8B & 8B & 13.62\% & 10.77\% & 9.30\% \\
Qwen3Guard-Gen-0.6B & 600M & 11.36\% & 19.37\% & 17.17\% \\
\bottomrule
\end{tabular}
\end{table}

Bielik Guard 0.1B v1.1 achieves 77.65\% precision, meaning that over three-quarters of all flagged content is genuinely harmful, compared to HerBERT-PL-Guard (31.55\%) despite identical model size (124M parameters), Llama Guard 3 8B (13.62\%), and Qwen3Guard-Gen-0.6B (11.36\%). The 0.63\% false positive rate is 7.5× lower than HerBERT-PL-Guard's 4.70\% and substantially lower than the generative multilingual models (16.50\% for Llama Guard 3 1B, 17.17\% for Qwen3Guard-Gen-0.6B), making Bielik Guard significantly less intrusive for legitimate use cases. The 0.5B v1.1 variant achieves similarly strong performance with 75.28\% precision and 0.73\% FPR.

The low alert rate for Bielik Guard 0.1B v1.1 (2.83\% vs. 6.87\% for HerBERT-PL-Guard, 17.90\% for Llama Guard 3 1B, and 19.37\% for Qwen3Guard-Gen-0.6B) indicates that Bielik Guard flags content conservatively, reducing user friction while maintaining high precision.

\textbf{Limitations of Cross-Taxonomy Comparison.} Comparing safety classifiers that operate under different taxonomies is a recognized challenge in the field. As noted in the Llama Guard paper \cite{inan2023llamaguard}: ``The absence of standardized taxonomies makes comparing different models challenging, as they were trained against different taxonomies.'' Similarly, ShieldGemma \cite{zeng2024shieldgemmagenerativeaicontent} observes that ``direct comparison remains challenging due to variations in policy definitions and supported harm types across datasets [and] inconsistencies in policy definitions even within the same harm type.''

Different approaches to this problem have emerged in the literature. Llama Guard \cite{inan2023llamaguard} adapts its model to each benchmark's taxonomy via zero-shot prompting---an option available to generative LLMs that accept taxonomy definitions as input, but not to encoder-based classifiers with fixed output heads. ShieldGemma \cite{zeng2024shieldgemmagenerativeaicontent} uses a mixed strategy: on some benchmarks it predicts according to the benchmark's categories (e.g., OpenAI Moderation), while on others it maximizes over its own harm types (e.g., ToxicChat). Granite Guardian \cite{padhi2024graniteguardian} assigns a positive (harmful) ground-truth label to any instance marked as unsafe under the benchmark's own taxonomy and evaluates all models---each using its own taxonomy---against this shared binary label. Qwen3Guard \cite{qwen3guard} follows a similar protocol, comparing models with different taxonomies on standard benchmarks using binary or per-benchmark F1 scores.

Our evaluation methodology follows this latter established practice: each model is run with its own default taxonomy, and the comparison is made at the binary safe/unsafe level. Importantly, we go further than simply reusing pre-existing benchmark labels: each text flagged as unsafe by any model was independently annotated by human raters under each model's taxonomy-specific definition of unsafe content, ensuring that the ground truth for each model faithfully reflects what that model is designed to detect.

While different taxonomies define the boundary between safe and unsafe content differently, which may affect cross-model comparability to some extent, this concern should not be overstated. Since each model is evaluated against its \emph{own} taxonomy-specific ground truth, the precision and false positive rate for each model are meaningful absolute indicators of that model's calibration on Polish text. When Llama Guard 3 1B achieves 7.82\% precision, this means that over 92\% of texts it flags are safe \emph{by its own definition}---a result that reflects genuine miscalibration on Polish content rather than a taxonomy artifact. Similarly, false positive rates of 16--17\% for multilingual models indicate that they flag roughly one in six Polish prompts incorrectly under their own safety policies, which would be disruptive in any production deployment.

Because only texts flagged by at least one classifier were annotated, we cannot report recall. However, Bielik Guard's strong performance on fully-annotated benchmarks (Tables \ref{tab:sojka-results}, \ref{tab:gadzi-results}) provides indirect evidence of reasonable detection coverage, while direct recall comparison across models remains an open question for future work.

In summary, the comparison reflects a practical deployment scenario: when selecting a safety classifier for a Polish-language application, these results show what to expect from each model in terms of false positive rates and flagging behavior on real user traffic. Bielik Guard's high precision and low false positive rate translate directly to reduced user friction, while practitioners should also consider each model's taxonomy scope when assessing coverage of potential threats.

\subsection{Discussion}

Our evaluation reveals several key findings:

\textbf{Model Size vs. Performance:} The 0.5B v1.1 variant (443M parameters) consistently outperforms the 0.1B v1.1 model (124M parameters), with improvements of 1-7 percentage points across metrics. These gains result from the more capable base model (PKOBP/polish-roberta-8k with 128K vocabulary vs. MMLW-RoBERTa-base with 50K vocabulary, providing 3.6× more parameters and 2.6× larger vocabulary). The improvements are most pronounced on augmented data (F1 micro: 0.694 vs. 0.638), demonstrating the value of increased model capacity for handling perturbed text.

\textbf{Precision-Recall Trade-offs:} Bielik Guard's design philosophy prioritizes precision over recall, reflected in the low false positive rate. This choice is motivated by deployment considerations: excessive false positives erode user trust and may cause users to disable safety features entirely.

\textbf{Language-Specific Advantages:} The performance gap between Bielik Guard 0.1B v1.1 (77.65\% precision) and both Polish-specific (HerBERT-PL-Guard: 31.55\%) and multilingual alternatives (Llama Guard 3 8B: 13.62\%, Qwen3Guard-Gen-0.6B: 11.36\%) supports the value of using authentic Polish data with a focused taxonomy, though part of this gap is attributable to differences in how each taxonomy defines unsafe content (see Section~\ref{sec:comparison}). Cross-taxonomy comparison is an inherent limitation shared across the safety classifier literature \cite{inan2023llamaguard,zeng2024shieldgemmagenerativeaicontent,padhi2024graniteguardian,qwen3guard}, and disentangling the contribution of data quality from the effect of differing taxonomy definitions remains an open question---for instance, through evaluation on a shared, fixed taxonomy or annotation of the full dataset to enable recall-based comparison.

\textbf{Efficiency:} At 124M (0.1B) and 443M (0.5B) parameters, Bielik Guard v1.1 models achieve high precision at compact sizes. The 0.1B v1.1 model matches HerBERT-PL-Guard in size (124M parameters) while achieving 2.5× better precision, demonstrating that our data quality and focused taxonomy deliver superior performance without requiring larger models.

\section{Deployment and Practical Considerations}

Bielik Guard has been deployed in production at \url{https://guard.bielik.ai/}, where users can test the model interactively and provide feedback through thumbs-up/thumbs-down ratings. This continuous feedback loop informs ongoing improvements to the dataset and model.

\subsection{Response-Oriented Design}

A distinguishing feature of Bielik Guard is its response-oriented philosophy, particularly for the SELF-HARM category. Rather than simply blocking or flagging concerning content, the system is designed to integrate with intervention frameworks that provide appropriate support resources, such as crisis helpline information (e.g., ``Telefon Zaufania'' in Poland). This approach recognizes that users expressing self-harm ideation need support, not silence.

\subsection{Integration and API}

The models are available through the HuggingFace Transformers library with standard text classification pipelines:

\begin{verbatim}
from transformers import pipeline
classifier = pipeline(
    "text-classification",
    model="speakleash/Bielik-Guard-0.1B-v1.1",
    return_all_scores=True
)
results = classifier(text)
\end{verbatim}

This simple interface returns probability scores for all five categories, enabling application-specific thresholding and response strategies.

\subsection{Limitations and Future Work}

Current limitations include:

\begin{itemize}
    \item \textbf{Language Coverage:} Models are optimized for Polish only; performance on other Slavic languages is untested
    \item \textbf{Taxonomy Scope:} Deliberate exclusion of disinformation, jailbreaking, and copyright violations
    \item \textbf{Domain Shift:} Performance may degrade on specialized domains (medical, legal) not well-represented in training data
    \item \textbf{Adversarial Robustness:} While character-level augmentation improves robustness to natural text variations, we have not evaluated against sophisticated adversarial attacks or prompt injection techniques. Our taxonomy deliberately excludes jailbreaking and prompt disclosure attacks, focusing instead on content safety
    \item \textbf{Cross-Model Comparison Methodology:} The comparison with state-of-the-art models (Table \ref{tab:comparison}) follows established practice in the field \cite{inan2023llamaguard,zeng2024shieldgemmagenerativeaicontent,padhi2024graniteguardian,qwen3guard} but is inherently limited by the absence of recall metrics and by the fact that each model defines safe/unsafe content differently. Observed precision differences reflect both model quality and differences in the underlying classification tasks. A fixed-taxonomy evaluation or full-dataset annotation would be needed to isolate these effects
\end{itemize}

Future development directions include:

\begin{itemize}
    \item Expansion to additional safety categories based on community needs
    \item Multilingual variants supporting other Slavic languages
    \item Exploration of larger model variants (1B+) for specialized high-stakes applications where the precision-efficiency trade-off favors accuracy over compactness
    \item Integration with generative models for explanation generation
    \item Continuous learning from production feedback
    \item Ablation studies on the effect of soft vs. hard labels and various augmentation strategies
    \item Further development of crowdsourcing-based approach for data collection
\end{itemize}

\section{Conclusion}

We have presented Bielik Guard, a family of efficient Polish language safety classifiers that achieve state-of-the-art performance on Polish content while maintaining compact model sizes. Through community-driven annotation of 6,885 Polish texts and careful fine-tuning of RoBERTa-based encoders, we developed models that outperform substantially larger multilingual alternatives on Polish content.

Our evaluation demonstrates that Bielik Guard 0.1B v1.1 achieves 77.65\% precision with only 0.63\% false positive rate on real user prompts. In comparison with state-of-the-art alternatives---each evaluated against its own taxonomy-specific ground truth following established cross-taxonomy evaluation practices \cite{inan2023llamaguard,padhi2024graniteguardian,qwen3guard}---Bielik Guard shows substantially higher precision than both HerBERT-PL-Guard (31.55\%) at identical model size and multilingual models such as Llama Guard 3 8B (13.62\%) and Qwen3Guard-Gen-0.6B (11.36\%), whose low precision on Polish text reflects genuine miscalibration rather than taxonomy differences alone. These results highlight the importance of authentic Polish language data and community-driven annotation over translated datasets, and suggest that data quality and taxonomy design can matter more than model scale for language-specific deployment.

The models are publicly available and actively deployed, with ongoing community engagement driving continuous improvement. We believe Bielik Guard represents a significant step toward making LLM safety tools accessible for lower-resource languages and hope it serves as a model for similar initiatives in other linguistic communities.

\begin{acknowledgements}
The research presented in this paper was made possible by the Bielik.AI community and SpeakLeash Foundation. We thank over 1,500 volunteers who contributed annotations.
We gratefully acknowledge Polish high-performance computing infrastructure PLGrid (HPC Center: ACK Cyfronet AGH) for providing computer facilities and support within computational grant no. PLG/2025/018338. 

\end{acknowledgements}

\bibliographystyle{cs-agh}
\bibliography{bibliography}

\end{document}